\documentclass{article}
\usepackage[dvipsnames]{xcolor}
\usepackage{sigbovik2022_conference,times}
\PassOptionsToPackage{hyphens}{url}
\usepackage[hidelinks,backref]{hyperref}
\usepackage{url}
\usepackage{CJKutf8}
\usepackage{soul}
\usepackage{dirtytalk}
\usepackage{epigraph}
\usepackage[labelformat=simple]{subcaption}
\usepackage{xspace}
\usepackage{pifont}
\usepackage{amssymb}
\usepackage{booktabs}
\usepackage{tabularx}
\newcolumntype{C}{>{\centering\arraybackslash}X}
\usepackage{multirow}
\usepackage{listings}
\usepackage{amsmath}
\usepackage{graphicx}

\usepackage{tikz}
\usetikzlibrary{arrows.meta}

\usepackage{pgfplots}
\pgfplotsset{width=15cm, height=7cm,compat=1.8,grid style={gray,opacity=0.25}}

\definecolor{codegreen}{rgb}{0,0.6,0}
\definecolor{codegray}{rgb}{0.5,0.5,0.5}
\definecolor{codepurple}{rgb}{0.58,0,0.82}
\definecolor{backcolour}{rgb}{0.95,0.95,0.92}

\lstdefinestyle{pystyle}{
    backgroundcolor=\color{backcolour},   
    commentstyle=\color{codegray},
    keywordstyle=\color{codegreen},
    numberstyle=\tiny\color{codegray},
    stringstyle=\color{codepurple},
    basicstyle=\ttfamily\footnotesize,
    breakatwhitespace=false,         
    breaklines=true,                 
    captionpos=b,                    
    keepspaces=true,                 
    numbers=left,                    
    numbersep=5pt,                  
    showspaces=false,                
    showstringspaces=false,
    showtabs=false,                  
    tabsize=2
}

\newcommand{\numMegaWatts}{23\xspace}

\usepackage[capitalize]{cleveref}
\crefname{section}{Sec.}{Secs.}
\Crefname{section}{Section}{Sections}
\Crefname{table}{Table}{Tables}
\crefname{table}{Tab.}{Tabs.}

\title{
A \numMegaWatts MW data centre is all you need
}

\author{Samuel Albanie,
Dylan Campbell, 
Jo\~{a}o F. Henriques\\
Pensive Prophets for Profit\\
Shelfanger, United Kingdom
}

\iclrfinalcopy 

\begin{document}
\maketitle

\begin{abstract}

The field of machine learning has achieved striking progress in recent years, witnessing breakthrough results on language modelling, protein folding and nitpickingly fine-grained dog breed classification.
Some even succeeded at playing computer games and board games, a feat both of engineering and of setting their employers' expectations.
The central contribution of this work is to carefully examine whether this progress, and technology more broadly, can be expected to continue indefinitely.
Through a rigorous application of statistical theory and
failure to extrapolate beyond the training data,
we answer firmly in the \textit{negative} and provide details: technology will peak at 3:07 am (BST) on 20th July, 2032.
We then explore the implications of this finding,
discovering that individuals awake at this ungodly hour with access to a sufficiently powerful computer possess an opportunity for myriad forms of long-term
linguistic `lock in'.
All we need is a large~($\gg 1$W) data centre to seize this pivotal moment.
By setting our analogue alarm clocks, we propose a tractable algorithm%
\footnote{Please see our Github repo (\url{https://github.com/albanie/A-23MW-data-centre-is-all-you-need}) for a permanent notice that code will be coming soon.}
to ensure that, for the future of humanity, the
British
spelling of \textit{colour} becomes the default
spelling across more than 80\% of the global word processing software market.%
\footnote{We grudgingly acknowledge that \texttt{color} arguments in the matplotlib API should retain their u-free spelling for backwards compatibility.
We are not complete barbarians.}%
\end{abstract}

\epigraph{Where are they?}{Enrico Fermi, \emph{CVPR 2022, virtual poster \#65713}}

\section{Introduction} \label{sec:intro}

Accurate forecasts are valuable.
From domains spanning battle outcomes~\citep{babylonians1900} to precipitation nowcasting~\citep{ravuri2021skilful}, 
humans have looked to hepatomancy and overly-cheerful weather presenters to assess the fate of their empire and to determine whether they need an anorak for their afternoon dog walk. 
Perhaps no topic has garnered more interest among professional forecasters than the future trajectory of technology~\citep{lucian155,voltaire1752,bush1945we}.
However, future prediction is a difficult business, and the historical record of this discipline is somewhat patchy.\footnote{In addition to a widely publicised failure to predict the near-term feasible exploitation of atomic energy, \cite{rutherford1933} also failed to predict the global blue-black/white-gold dress debate of 2015.}

Science fiction authors have fared better at predicting the future, if you cherry-pick enough.\footnote{Even your first CIFAR-10 model was right 10\% of the time.}
Rockets land themselves, some cars drive themselves (when someone is looking), and some humans think for themselves (and others). The opposing visions of \cite{orwell} and \cite{huxley1932brave} predicted two dystopias, one where people were controlled by fear and surveillance, another where they were controlled by endless entertainment and distraction.
Rather than assess their fidelity, let us move swiftly on.
\cite{foundation} proposed psychohistory as the science of predicting the future behaviour of human populations.
By analogy to particles in a gas---assuming perfectly spherical humans in a vacuum---while each human's state is unknowable, one can derive statistical quantities such as peer pressure, shear stress, and cringe factor.

To take on the challenge of future technology prediction, several themes have emerged in prior work.
One body of research has leveraged historical trends and hardware laws with hints of exponential nominative determinism to underpin forecasts~\citep{kurzweil1990age},
lightly handcuffed by physical limits~\citep{bennett1985fundamental}.
A second approach, pioneered by~\cite{gabor1963inventing}, acknowledges the impossibility of future prediction and instead advocates inventing the future, or sagely producing smarter agents to take care of the inventing~\citep{yudkowsky1996staring}. 
However, empirical scaling laws lack a principled theoretical foundation, while actively inventing the future sounds exhausting, and honestly just waiting for Moore's law is much easier.

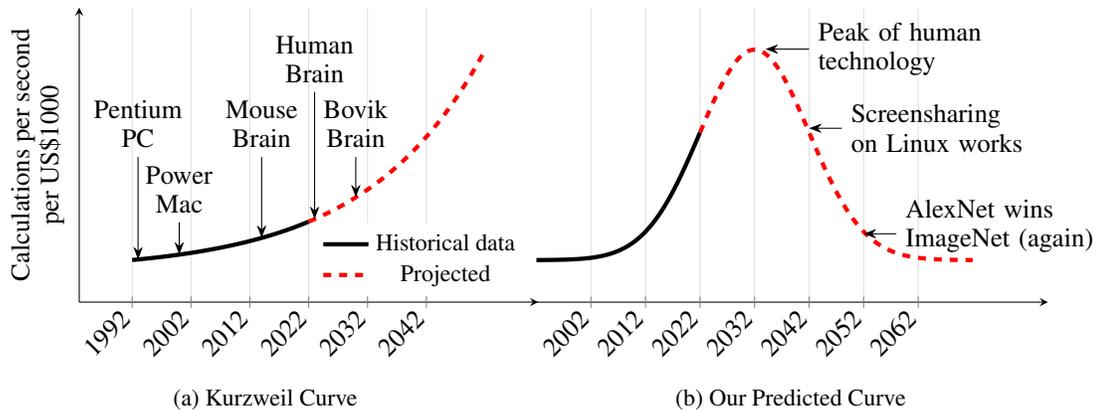
\begin{figure}\centering
    \begin{subfigure}{0.5\textwidth}
        \begin{tikzpicture}
\begin{axis}[
  width=1.1\textwidth,
  height=5.5cm,
  legend style={font=\small,anchor=south east, at={(0.98,0.02)}, fill=white, align=left, draw=none},
  axis x line=bottom, %
  axis y line=left,
  extra x ticks={-2, -1.5,-1,-0.5,0,0.5},
  extra x tick labels={1992, 2002, 2012, 2022, 2032, 2042},
  extra tick style={grid=major, tick label style={rotate=45,anchor=east}},
  xtick=\empty,
  ytick=\empty,
  ylabel style={align=center},
  ylabel=Calculations per second \\ per US\$1000,
  enlarge x limits=0.15,
  enlarge y limits={0.2},
  ]
  \addplot[domain=-2:-0.5,mark=none,samples=50,smooth,style=ultra thick] {exp(x)};
  \addplot+[domain=-0.5:1.0, mark=none, dashed,samples=50,smooth,style=ultra thick] {exp(x)};
  \legend{Historical data, Projected}
  \draw [{Stealth}-](axis cs:-1.95,{exp(-1.95)}) -- (axis cs:-1.95,1.4) node [above,text width=1.5cm,align=center] {Pentium PC};
  \draw [{Stealth}-](axis cs:-1.6,{exp(-1.6)}) -- (axis cs:-1.6,0.6) node [above,text width=1.5cm,align=center] {Power Mac};
  \draw [{Stealth}-](axis cs:-0.9,{exp(-0.9)}) -- (axis cs:-0.9,1.4) node [above,text width=1.5cm,align=center] {Mouse Brain};
  \draw [{Stealth}-](axis cs:-0.45,{exp(-0.45)}) -- (axis cs:-0.45,2.2) node [above,text width=1.5cm,align=center] {Human Brain};
  \draw [{Stealth}-](axis cs:-0.1,{exp(-0.1)}) -- (axis cs:-0.1,1.4) node [above,text width=1.5cm,align=center] {Bovik Brain};
\end{axis}
\end{tikzpicture}%
        \caption{Kurzweil Curve}%
        \label{fig:kurzweil_curve}
    \end{subfigure}\hfill
    \begin{subfigure}{0.5\textwidth}\centering
        \pgfmathdeclarefunction{gauss}{2}{%
  \pgfmathparse{1/(#2*sqrt(2*pi))*exp(-((x-#1)^2)/(2*#2^2))}%
}

\begin{tikzpicture}
\begin{axis}[
  width=1.2\textwidth,
  height=5.5cm,
  clip mode=individual,
  axis x line=bottom, %
  axis y line=none,
  extra x ticks={-1.5,-1,-0.5,0,0.5,1.0,1.5},
  extra x tick labels={2002, 2012, 2022, 2032, 2042, 2052, 2062},
  extra tick style={grid=major, tick label style={rotate=45,anchor=east}},
  xtick=\empty,
  ytick=\empty,
  ylabel style={align=center},
  enlarge x limits={abs value=1cm,upper},
  enlarge y limits={0.2},
  ]
  \addplot[domain=-2:-0.5,mark=none,samples=50,smooth,style=ultra thick] {gauss(0,0.5)};
  \addplot+[domain=-0.5:2, mark=none, dashed,samples=50,smooth,style=ultra thick] {gauss(0,0.5)};
  \draw [{Stealth}-](axis cs:0.1,0.8) -- (axis cs:0.5,0.8) node [right,text width=2.5cm,align=left] {Peak of human\\technology};
  \draw [{Stealth}-](axis cs:0.5,0.5) -- (axis cs:0.8,0.5) node [right,text width=3cm,align=left] {Screensharing\\on Linux works};
  \draw [{Stealth}-](axis cs:1.0,0.1) -- (axis cs:1.3,0.1) node [right,text width=2.5cm,align=left,yshift=3pt] {AlexNet wins ImageNet (again)};
\end{axis}
\end{tikzpicture}%
        \caption{Our Predicted Curve}%
        \label{fig:our_curve}
    \end{subfigure}
    \caption{\textbf{A principled approach to future extrapolation.}
    Standard approaches to future prediction, exemplified by the curve of~\cite{kurzweil2005singularity}, are guided by empirical data and hardware trends. %
    By contrast, our prediction relies on the tried and tested Central Limit Theorem.
    Note how ours is more symmetric and visually appealing.
    We note that the vertical\protect\footnotemark \,axis breaks down in the post-2032 regime, where US\$1000 is increasingly meaningless.
    We therefore convert to the equivalent amount of pickled herring, and proceed.
    We will also forecast several events using our predictive model, which we now predict will be detailed in a later section of this article.
    }
    \label{fig:future_curves}
\end{figure}
\footnotetext{According to Wikipedia, ``the word `vertical' corresponds to a graph as traditionally rendered on graph paper, where vertical is oriented toward the top of the page, regardless of whether the page itself---or screen in the computer era---is embedded upright or horizontally in physical 3-space. The `top' of a page is itself a metaphor subordinate to the convention of text direction within the writing system employed. `Vertical' has a concrete as opposed to metaphorical meaning within gravitational frames; when a page is held `upright' in 3-space these two concepts align with the top of the page also gravitationally vertical. Horizontal is often equated with `left' and `right', but note that in the typographic convention of recto and verso, left and right also take on additional meanings of front and back'' \citep{wiki2022abscissa}.}

In this work we propose a third approach that both benefits from rigorous statistical theory and has a higher chance of completion within the modern 37.5 hour UK working week (including tea breaks).
Our starting point was to turn to that reliable workhorse of modern statistical theory, the Central Limit Theorem.
In brief, the Central Limit Theorem states that when random variables with sufficient independence of thought are summed, their normalised summation tends asymptotically towards a Gaussian distribution. 
We must, of course, address the delicate question of whether the Central Limit Theorem can legitimately be applied to our future forecasting problem.
Thankfully, the Central Limit Theorem can and should be applied to all problems involving uncertainty, and few topics are as uncertain as the future.\footnote{Historical misapplications of the Central Limit Theorem have arisen simply by applying the wrong variant of the Central Limit Theorem---there are a wonderful assortment of variants to choose from.
Out of respect for the venerated theorem, we never acronymise.}

The technical foundations thus laid, the first key contribution of this work is to observe that recent decades of exponential growth in the number of transistors lovingly squeezed onto a microchip\footnote{Some of which can be attributed to Tesco's convenient part-baked microchips, which can be finished in a home oven to a golden crisp.} neatly fits the steep climb of a hand-drawn bell curve (\cref{fig:future_curves}).
A textbook application of the Central Limit Theorem then yields two striking insights. 
First, it enables us to compute with extremely high precision the date at which technological progress (as measured by transistor snugness or FLOPs per 1K USD) will peak: \textit{3:07 am (BST) on 20th July, 2032}.
Second, it enables dead certain modelling of uncertainty in our predictions, because the prediction itself is a Gaussian distribution (note that we carefully select our technology units such that the area under the curve sums to one).
With the future now standing a little more naked in front of us, it behooves us to consider the questions motivated by this discovery, discussed next.

\textit{What is the cause of the decline?}
While prior work has explored the stasis-inducing potential of soma~\citep{huxley1932brave} and drouds~\citep{niven1969death}, we initially posited that widespread use of large language models for coding and paper writing will result in increasingly automated science production, with the associated atrophy of biological brains, and a drift of technological focus to matters of interest to disembodied machines.%
\footnote{These include beating all other models on larger boards of multi-coloured Go, achieving 100\% accuracy on MNIST, and a sequence of increasingly sophisticated puns about endianness.}
However, our later findings suggest that a simple coupling of reward hacking with an ageing Flappy Bird clone will bring about the reversal.

\textit{What research opportunities does the decline present?}
Options abound for machine learning researchers who can acquire control of a \numMegaWatts MW computer at 3:07 am.
While some may choose to lay down an ImageNet top-1 SoTA that outlives Bradman's test career batting average, we propose instead to pursue a worthier cause.
Leveraging a novel algorithm (described in \cref{sec:spelling}), we plan to conduct a 51\% attack on spellings of the word \textit{colour} amongst the internet corpora used to train autocorrect systems.
By doing so, we protect ourselves from annoying red-underlined squiggles well into our retirement, or worse, learning several new spellings.

\textit{Will we have to give up wireless headphones?}
Sadly.
According to curve, by 4000 AD the last of the great homo sapiens engineers will strain to build an Antikythera orrery.
It will probably be flat.

The remainder of this paper is structured as follows.
In \cref{sec:related}, we wilfully misrepresent prior work to increase our chances of acceptance at the prestigious SIGBOVIK conference. 
Next, in \cref{sec:method}, we describe in more detail our mischievous methodology to topple the status quo among the spelling tsars.
Finally, after minimal experimentation in \cref{sec:experiments}, we conclude with conclusions in \cref{sec:conclusion}.

\epigraph{Prediction is very difficult, especially if it's about the future.}{Niels Bohr} 

\section{Uncomfortably Closely Related Work \label{sec:related}}

\textbf{Quo vadis? A short history of future prediction.}
Formative early work by~\cite{delphi} at Delphi showed the considerable value of goat sacrifice, noxious fumes and keeping just a hint of ambiguity when forecasting (revived recently as ``confidence intervals'').
In the more recent machine learning literature, creative researchers have sought to learn representations by predicting the near future in video~\citep{ranzato2014video,vondrick2016anticipating}.
To the best of our knowledge, however, no such work has considered video extrapolation decades into the future, which would clearly be more useful.
More related to our time scale, we note that the farsighted ``standard run'' \textit{limits to growth} model of~\cite{meadows1972limits} for population growth adopts a similar ``what goes up must come down'' philosophical bent to our forecast, but their graph is spiritually closer to a Laplace distribution than a Gaussian
and hence the Central Limit Theorem cannot be so confidently invoked.

\textbf{Claims of universal sufficiency.}
Other than our own, the most notable past attempts to make gloriously broad claims about a framework's ability to fulfil all of a researcher's needs have considered Trust~\citep{welter2012all}, A Good Init~\citep{mishkin2016all}, Attention~\citep{vaswani2017attention} and Love~\citep{beatles1967}.
We note with consternation that, as a corollary of being all-encompassing, the above must be mutually exclusive.
This unfortunate property is a product of clumsy first-order logic, and may be relaxed in the future by a reformulation using cold fuzzy logic, and warm fuzzy feelings.
A cautionary note: not all prior literature has your best interests at heart.
Previous work has attempted to claim ``all your base''~\citep{zerowing1991} and employed the elegant tools of algebraic geometry to construct $n$-dimensional polytope schemes that efficiently separate you from your financial savings~\citep{fouheyn}.

\epigraph{Who controls the past controls the future: who controls reddit spelling convention controls the past.}{George Orwell, \emph{1984}}

\section{The road ahead \label{sec:method}}

As noted with some trepidation in the introduction, the Central Limit Theorem assures us that the technological progress of humanity will soon stall before
relinquishing its historical gains.
In this section, we first explore 
possible causes of this trajectory (\cref{subsec:causes}).
We then fearfully investigate the implications of our findings (\cref{subsec:implications}).

\subsection{Causes} \label{subsec:causes}

We initiated our analysis of plausible causes of technological decline by enumeration:
a butterfly wing flap in downtown Tirana in 1658;
Poincar\'e's premature death in 1912;  
the inexplicable density of Milton Keynes' roundabouts in 2022.
Yet none of these seemed fully adequate.

The science fiction literary canon suggests an alternative cause. %
In desperate search of relief from Slack notifications, increasing numbers of technologists over the next decade will turn to 
\textit{soma}~\citep{huxley1932brave} and reasonably priced wireheading options~\citep{niven1969death}, thereby functionally removing themselves from the innovation collective.
However, although it is reasonable to expect a one-way loss of many great minds to these attractors (slowing the progress curve), our modelling suggests that a significant fraction of the engineering community have already submitted to the iron rule of \textit{focus mode}. %
These hardy individuals 
are likely immune to such sirens, though they are hungry, owing to their missing two out of every three food deliveries and three out of three impromptu group trips to Nando's.
We therefore sought to understand how the last bastion of resilient \textit{focus moders} too could fall.
To this end, we trained a one-and-a-half layer Transfomer~\citep{vaswani2017attention}
to predict future events from a series of carefully curated, temporally ordered Wikipedia entries.

By sampling from the transformer, we learnt that by the year 2031, humanity will have achieved a new \textit{Gini coefficient} SoTA, wisely distributing 99.99999\% of its total wealth among three living humans (each unreachable by Slack notifications) and an adorable Labrador named Molly.
It is into such a society that on December 31st, 2031, an anonymous game designer\footnote{We can't be sure who. But if we had to guess, it would be SIGBOVIK legend, Dr Tom Murphy VII.} %
releases an adaptive, self-learning reboot of the 2013 mobile classic, \textit{Flappy Bird} (FB).
Designed to maximise user engagement, the FB algorithm soon begins to explore reward hacking strategies.
Although it garners a following of just 17 users, its impact is monumental.
Of these 17, three sit atop the Forbes ``3 over 30 trillion USD'' list, and each is incapacitated, unable to stop playing lest they lose their progress (which cannot be saved). 
Shortly thereafter, global capital flows grind to a halt and the silicon sandwich factories begin to fall silent.

In shock, the world turns to Molly.
She would like to help if she could, but only after walkies, and she's not sure that she can remember her BTC passwoof [sic].
Starved of donations, Wikipedia servers are powered down, and the sole Stack Overflow answer explaining how to undo recent git commits is lost.
Alas, any replications of this sacred knowledge were deleted long ago as duplicates by ModeratorMouse64.
A period of mourning ensues.
There is still hope. 
The situation could be salvaged.
But it requires the technologists to speak to other humans.
And so, because that would be awkward, the opportunity is lost.

\epigraph{Perfectly balanced, as all things should be.}{ImageNet-1K (2009)}

\subsection{Implications} \label{subsec:implications}

We next consider implications.
We begin by noting that modern ``brain-scale'' language models represent not only a great demo and a potential existential threat to humanity: they also open up a clear attack surface on the previously impregnable English spelling and grammar kingdom.
The reason is simple.
Just as doing away with cruft is a programmer's biggest joy, we ditch old headache-inducing paradigms with enthusiasm.
As we transition from fast approximate string matching to language models all the way down, the battleground of canonical spelling moves from carefully curated small-scale corpora to vast, scarcely filtered swathes of the internet.

To exploit this observation, we propose an approach inspired by the 2007 run on the UK bank, Northern Rock, and its exemplary demonstration
of a \textit{positive feedback loop}.
Noting that members of the celebrity GPT family of models~\citep{radford2018improving,radford2019language,brown2020language} are trained via log-likelihood maximisation, our objective is to ensure that 51\% of the training data adopts our preferred spelling conventions.
To operationalise this plan, we propose
\textsc{HMGAN} (Her Majesty's Generative Adversarial Network),
a neural network architecture inspired by~\cite{goodfellow2014generative} that ingests sentences of written English and rephrases text to maximise a measure of similarity against a corpus of text respectfully inspired by blog posts penned by Queen Elizabeth II. 
Since future auto-correcting spelling assistants will likely derive from future generations of these models, and since human authors passionately hate both red squiggly lines beneath words and changing default settings, a simple majority in frequency counts across training corpora should suffice to ensure that future text converges asymptotically to our convention.

\subsection{This spells trouble} \label{sec:spelling}

In the tradition of machine learning papers, we have many biases and assumptions that underpin our choice of datasets and targets. Against tradition, we list some of ours here. Our canonical English has the following features.
\begin{enumerate}
    \item U-philic: wherever there is an ``or'', there ought to be an ``our'', except where that isn't true. We support colour, valour, and flavour, but disown doour, wourk, and terrour.\footnote{No wonder we need an AI spell-checker, clearly these rules are not specifiable. That being said, two out of two linguists we interviewed suggested that computer scientists shouldn't be deciding these things.}
    \item Zed-phobic: zed (or izzard) is supposed to be a shock.\footnote{Oxford disagrees, and has a well-publicised infatuation with z that beggars belief. It is a rare beggar that believes an etymological $\zeta$ trumps a curvy French \textit{s}.} Incurs a penalty in our loss function.
    \item Ell-shippers: ells are meant to be together. It would be an unethical AI practice to separate them at this time. %
    \item Singular variants for all \textit{pluralia tantum}: a trouser, a pant, a scissor, and a thank are all valid choices under our orthography.
    \item Tildes: allowable in mathematics, and in \~{a} (the authors declare no conflict of interest in this decision). %
\end{enumerate}
We seek to provide the tools for baking in the consensus mode, which we will be releasing open source, with the stipulations that they not be used by anyone seeking to promote `color' over `colour' or by \textit{les immortels} of the Acad\'{e}mie Fran\c{c}aise%
\footnote{The astute reader may note that we nevertheless cherish our French linguistic influences, favouring the Anglo-French \textit{colour} over the Latin \textit{color}.
}.

\subsection{A critical point} \label{sec:compute_power}

To plan ahead, as required by our grant application,
we must address two key questions.
First, what magnitude of computing muscle is required to conduct a successful 51\% spelling attack on the global internet?
Second, how can we ensure that the spell checker trained post-attack achieves and maintains pole spell-check position on \textit{paperswithcode}, thereby ensuring uptake and the positive feedback loop we seek to create?

\noindent \textbf{Charting the future compute curve.}
Forecasting future computation is fraught with difficulty, but forecasting the power draw of future computers may be simpler, and thus we turn to this approach.
Over the past decade, efficiency gains have ensured that increases in energy consumption across global data centres have been fairly modest, growing approximately 6\% between 2010 and 2018, reaching 205 TWh~\citep{masanet2020recalibrating}.
Through a combination of tea-leaf interpretation and curve fitting, we estimate that global data centre energy usage will be in the region of 227 TWh in 2032.

Before turning to the implications of this estimate, let us note a few additional salient observations.
First, thanks to healthy market competition in the services sector, it is likely that an increasing fraction of the world's computing budget will be allocated to the operation of friendly sales chatbots.
By 2032, we believe that almost all written English content appearing on the internet will arise in unintentional bot-to-bot conversation.
As such, the training corpora for future spell checkers will be curated almost entirely from their transcripts (after filtering out the phrase ``I'm sorry, I didn't understand that. Did you mean `How can I give you five stars on Amazon?''').
Second, note that approximately 0.01\% of written English~\citep{leech1992100,davies2010corpus} corresponds to usage of the word `colour` or `color'---a frequency that we assume will be reflected in the chatbot discourse.
Third, observe that the best spell checkers must keep themselves relevant by training on only the most recent linguistic usage (discussed in more detail below). 

In light of the above, we see that a successful 51\% attack to establish a simple majority spelling of `colour' can be achieved by surpassing global chatbot text generation for a short period of time---just long enough for spell-checkers to fixate on the new spelling.
By employing a text synthesis algorithm (HMGAN) whose energy consumption matches that used by chatbots, we find that a \numMegaWatts MW data centre suffices for our aims 
(a derivation of this estimate can be found in~\cref{app:power-estimates}).
Since the chatbots will, of course, rely on the latest spell-checkers to avoid embarrassing their corporate overlords, they will quickly transition to the new spelling.
Then, as technology begins to decline, content production will drop, and spell-checkers will be forced to consider ever-expanding temporal windows to curate sufficient training data, rendering it ever more costly to reverse the trend.

\noindent \textbf{A timeless spell-checker.}
If spell checkers are to keep up to date with modern argot (similar to, but decidedly not, a fungal disease), it is critical that they are trained on the most recent and most correct data. To this end, we propose a diachronic language model spell-checker. Extending the work of \citet{loureiro2022timelms}\footnote{An admirable instance of the ``lour'' convention.}
to meet our needs, we commit to releasing a language model and spell-checker update every three hours, trained on a carefully curated and corrected stream of Twitter data. Our last update, for example, was trained on 123.86 million tweets, weighted according to the logarithm of the number of retweets and hearts, and with spelling errors corrected where appropriate. Importantly, our time-aware language model has knowledge of recent events and trends, allowing us to capture language as it is used in practise, not how the Acad\'{e}mie Fran\c{c}aise ordains.
For example, we observed a significant spike in the incidence of five-letter words, especially those with many vowels.
Unlike existing language models, ours was successfully able to mirror this trend and dilate or contract words entered by our users to five letters. An unforeseen side-effect was the conversion of some words to coloured rectangles
{$\color{Gray}\blacksquare \color{OliveGreen}\blacksquare \color{Yellow}\blacksquare \color{Gray}\blacksquare \color{Yellow}\blacksquare$},
but this is likely a consequence of our data augmentation strategy.
It is crucial that all language-based tools be kept abreast of recent events and trends, because AI models of this sort deep freeze the cultural landscape from where the training data is obtained. It is highly unethical for AI researchers to participate in a system that creates cultural feedback loops and stagnation, over-privileging the status quo at the expense of the kids\footnote{The spelling of colour is the only exception to this rule.}. We further observe that Twitter is an excellent and unbiased source of international language usage that does not reflect any one cultural background, and so is a particularly good dataset for our purposes. It is also on the Acad\'{e}mie's list of banned linguistic sources, which in our view speaks to its merits.

However, it is not enough to periodically release a language model fine-tuned on the last 3 hours of corrected Twitter data. In the fast-evolving world of language, this is already unusably out of date. Our previously-described model failed, for example, to autocorrect ``vacation'' to ``staycation''.
It is incumbent on GPT-as-a-service (GaaS) providers to provide up-to-the-minute language models, motivating the development of temporally-predictive models.
As we shall show in our experiments, our Predictive Diachronic Generative Pretrained Transformer (PDGPT) model effectively captures contemporary language usage, reflecting the most recent events, and is moreover able to generate geo- and time-localised results.

\epigraph{You either die a grad student, or you live long enough to become R2.}%
{Dr. Harvey Dent (NeurIPS Area Chair), 2008
}

\section{Experiments} \label{sec:experiments}

In this section, we first validate our ideas in a simplified setting by considering 51\% attacks in the context of the \textit{British Bin Colouring Problem} (\cref{sec:bin_colouring}).
We then compare our PDGPT spell-checker to the existing state-of-the-art (\cref{sec:sota}) and discuss civilisational impact (\cref{sec:broader-impact}).

\subsection{The British Bin Colouring Problem}
\label{sec:bin_colouring}

The \textit{British Bin Colouring Problem} (BBCP) refers to a mathematical problem that is more practical than graph colouring and more general than bin packing.
The task is as follows.
On Wednesday evenings (or your local bin collection night), the objective is to wheel out the colour of bin that causes maximum mischief to your neighbours.
Wheeled out bins of the wrong colour \textit{will not be collected under any circumstances}.
You have three choices:
(1) black - unfiltered,
(2) blue - recycling,
(3) green - garden waste.
Central to this problem is the assumption that, to avoid social tension, almost all neighbours will copy their neighbours' bin colour, rather than check the official bin collection colour through the local government website.
Note, that you must account for upstanding citizens, who will put out the right bin colour regardless of their neighbours, misleading leaflets, or inaccurate local government websites.
The problem is NP-Hard and environmentally significant.

We consider an instance of the BBCP for the residents of Grantchester, a picturesque village in Cambridgeshire.
Our strategy was simple: we first employed HMGAN to craft a sequence of royal entreatments to wheel out the blue coloured bin on a green bin Wednesday, and sent leaflets to this effect at addresses generated via a Sobol sequence to ensure reasonable coverage.
We then wheeled out our own blue bin and waited.
A combination of stochastic upstanding citizen locations and wheel-out race conditions complicated our analysis, leaving us in some doubt as to whether a 51\% bin colour majority would achieve our desired ends.
To counter this intractability, we employed a systematic strategy of hoping it would work.

Unfortunately, the results of this experiment were unpromising.
In our enthusiasm, we had failed to wait until 27th March, thereby missing the transition to Daylight saving time.
As a consequence, it was too dark for our neighbours to determine our bin colour and were thus uninfluenced.
They also did not take kindly to unsolicited leaflets, and are, by now, quite frankly tired of our shenanigans.

\subsection{Simulated comparison to the state-of-the-art}
\label{sec:sota}

\begin{figure}
    \centering
    \includegraphics[width=0.7\textwidth]{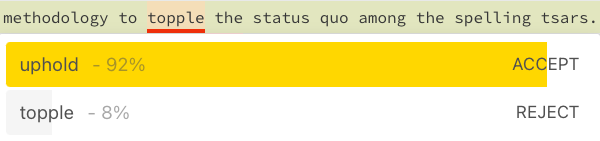}
    \caption{\textbf{When it comes to spelling, it's not so easy to topple the status quo.}
    Increasingly dystopian modern grammar checkers, when applied to the close of the introduction of this article, let us know that we stand little chance of success.
    We soldier on.}
    \label{fig:spelling}
\end{figure}

Undeterred, we turn next to an evaluation of our PDGPT spell checker, capable of both autocorrection and event prediction.
By backtesting on historical data, we find events and spellings successfully predicted or caused by our model include
\textit{quarantinis} but not \textit{maskne}.
More concerningly, despite our comprehensive set of three unit tests, PDGPT insists on auto-correcting our own use of `colour' to `color', undermining the core objective of our enterprise.
This speaks to the formidable challenge of over-turning the spelling status quo (see \cref{fig:spelling}), the difficulty of controlling large language models and the fact that we still don't really understand what the \texttt{.detach()} function does in PyTorch.

\begin{table}[t]%
	\caption{Masked token prediction for our Predictive Diachronic Generative Pretrained Transformer (PDGPT). For each three-hourly model, the table displays the top-3 predictions ranked by their prediction probability.}
	\label{tab:prediction}
	\begin{tabularx}{\linewidth}{l C C C}
		\toprule
		Models for & I'm working & I keep forgetting & Looking forward to\\
		01/04/2022 & from $\langle$mask$\rangle$. & to bring a $\langle$mask$\rangle$. & watching $\langle$mask$\rangle$!\\
		\midrule
		\multirow{3}{*}{09:00 UTC} & bed & smile & closely\\
		& home & purpose & yall\\
		& afar & baguette & snow\\
		\midrule
		\multirow{3}{*}{12:00 UTC} & home & bag & snow\\
		& upstairs & mug & skaters\\
		& tenerife & charger & twitch\\
		\midrule
		\multirow{3}{*}{15:00 UTC} & memory & charger & tv\\
		& home & friend & bridgerton\\
		& work & bottle & ash\\
		\midrule
		\multirow{3}{*}{18:00 UTC} & shelter & torch & flames\\
		& cover & bottle & revelry\\
		& asgard & party-hat & ragnarok\\
		\bottomrule
	\end{tabularx}
\end{table}

In \cref{tab:prediction}, we present qualitative results from our three-hourly predictive models trained for 01/04/2022\footnote{We presume our model uses the DD/MM/YYYY convention.}. Our model predicts the $\langle$mask$\rangle$ token in context, the same mode we use for text auto-completion. While we are not yet able to evaluate the quality of these predictions, we expect them to be rigorously validated by the time of publication. We note that our model has learned to reason about localised weather systems, plausibly predicting snow late in the season with no actual meteorologically-relevant input.

\subsection{Limitations, Risks and Civilisational Impact}
\label{sec:broader-impact}

One limitation of our approach is reflected in our complete inability to produce convincing experimental results to date, even in Grantchester. 
We believe that this limitation will be overlooked by reviewers who recognise other merits to our work, such as our heavy use of footnotes which lend much needed academic gravitas to the text.

A risk of our approach is that it may encourage other researchers, notably our beloved American colleagues, to pursue a similar framework, escalating into a transatlantic arms race in which ever larger fractions of the planet's energy are dedicated to controlling spelling conventions.

In terms of civilisational impact, the stakes are as high as ever.
John Wesley, the founder of Methodism, notably considered the removal of the \textit{u} a `fashionable impropriety' in 1791~\citep{mencken1923american}.
But in 2032, for the first time the opportunity will exist for eternal spelling lock in  for the large swathe individuals who don't remember to change the default setting on their spell-checker.

\section{Conclusion} \label{sec:conclusion}

We have presented a rigorous statistical analysis of historical and future computational trends and ascertain the date and time at which technology, on average, will peak. We have leavened this potential downer with an account of the implications of this finding and the concomitant opportunities that this presents. We provide the tools for myriad forms of long-term cultural and linguistic ``lock in'', with a particular focus on spelling and an especial concern for that of ``colour''. We expect this work and attitude to resonate throughout the following crepuscular decades as we revert to our respective agrarian utopias.

There are many promising avenues for future research. However, the relatively brief time before technological decline sets in precludes large-scale projects if significant computation is required by the work.
One correspondent has suggested that a mixture-of-Gaussians model is more appropriate for our extrapolation, to better conform to conformal cyclic cosmology \citep{penrose2010cycles}, as all theories must. A mixture-of-infinite-Gaussians is intellectually appealing, but computationally infeasible (without using a RKHS, which is unfashionable).
A more plausible direction is a new time and date scheme based on standard deviations from the Gaussian technology curve.
Significant further research is required to quantify whether the two-to-one mapping from years to standard deviations will be problematic.

\noindent \textbf{Acknowledgements.}
We acknowledge with gratitude our families, friends, colleagues, and cheerful international research community.
We also acknowledge the profound wisdom of A. Sophia Koepke.

\bibliographystyle{iclr_style}
\bibliography{refs.bib}

\appendix

\newpage

\section{Compute power estimates}
\label{app:power-estimates}

In this appendix, we provide detailed calculations of our data centre size requirements.
In the spirit of appendices, which are not written to be read, the derivation was carefully not checked for errors.
Computations were performed on an iPhone with a slightly cracked screen.

\cite{masanet2020recalibrating} estimated global data centre energy consumption in 2018 at 205 TWh, corresponding to a rise of 6\% since 2010.
Extrapolation over 14 years at a growth of 6\% every eight years yields an estimate of 227.007 TWh for 2032, or 25.914 GW average power.
The present edition of the Corpus of Contemporary American English~\citep{davies2010corpus} indexed by~\cite{englishcorpora}, finds 124814 occurrences of `color' among 1 billion words (constituting 0.0124814\% of all word usage).
The British National Corpus~\cite{leech1992100} contains 11,135 counts of `colour' among 100 million words (amounting to 0.011135\% of all word usage).
Averaging these terms, we determine that 0.0118082\% of written English (at least, the kind that makes it into linguistic corpora) consists of some variant of the word colour.
By 2032, we assert that all but a vanishing fraction of written text will consist of conversation between chatbots, which will dominate power consumption. 
Approximately 60.6\% of internet content is written in English at present~\citep{w3techs}, and we assume the chatbots will follow this trend.
We further assume, naturally, that the chatbots will use spell checkers, and will approximately maintain historical word frequency usage.
Thus, it becomes clear that a successful 51\% attack requires us to obtain a proportional share (by power) of global data centre compute that will outpace the use of colour in chatbot-to-chatbot conversation.
This amounts to $1.91147$ MW = ($26.712328 \mathrm{GW} \times 0.606 \times 0.0118082 / 100) \times 1000$.
Thanks to the Matthew law, and the fact that spell checkers will be trained only on the most recent corpora to avoid staleness, we need only overwhelm the text generation of the chatbots for a very short period of time.
However, note that our attack strategy requires us to insert instances of `colour' into minimally valid sentences that will make it into the spell checker training data. 
On believe that pithy sentences with an average length of 11 words will suffice for this task, 
bringing our power needs for 22.93764 MW.
Finally, we allocate a safety buffer of 0.01 MW capacity to install involuntary operating system updates forced upon us at 3:06 am, leading to a 22.94764 MW power profile.

\end{document}